\documentclass[letterpaper, 10 pt, conference]{ieeeconf}  

\usepackage{adjustbox}
\usepackage{graphicx}
\usepackage{utfsym}
\usepackage{booktabs}
\usepackage{array}
\usepackage{cite}
\usepackage{algpseudocode}
\usepackage[linesnumbered,ruled,vlined]{algorithm2e}
\usepackage{xcolor}
\usepackage{multirow}

\newcommand{\Tau}{\mathrm{T}}

\IEEEoverridecommandlockouts                              

\title{\LARGE \bf
SFTrack: A Robust Scale and Motion Adaptive Algorithm\\for Tracking Small and Fast Moving Objects
}

\author{Inpyo Song$^{1}$ and Jangwon Lee$^{1}$
\thanks{$^{1}$Department of Immersive Media Engineering
\& Department of Computer Science Education, Sungkyunkwan University, Seoul, Korea
{\tt\small e-mails: Inpyo Song(songinpyo@skku.edu) \& Jangwon Lee(leejang@skku.edu)}}
}

\begin{document}

\maketitle
\thispagestyle{empty}
\pagestyle{empty}


\begin{abstract}
This paper addresses the problem of multi-object tracking in Unmanned Aerial Vehicle (UAV) footage.
It plays a critical role in various UAV applications, including traffic monitoring systems and real-time suspect tracking by the police. However, this task is highly challenging due to the fast motion of UAVs,
as well as the small size of target objects in the videos caused by the high-altitude and wide-angle views of drones.
In this study, we thus introduce a simple yet more effective method compared to previous work to overcome these challenges.
Our approach involves a new tracking strategy,
which initiates the tracking of target objects from low-confidence detections
commonly encountered in UAV application scenarios. 
Additionally, we propose revisiting traditional appearance-based matching algorithms
to improve the association of low-confidence detections.
To evaluate the effectiveness of our method,
we conducted benchmark evaluations on two UAV-specific datasets (VisDrone2019, UAVDT) and one general object tracking dataset (MOT17).
The results demonstrate that our approach surpasses current state-of-the-art methodologies,
highlighting its robustness and adaptability in diverse tracking environments.
Furthermore, we have improved the annotation of the UAVDT dataset by rectifying several errors
and addressing omissions found in the original annotations.
We will provide this refined version of the dataset to facilitate better benchmarking in the field.
\end{abstract}

\section{Introduction}
Multi-object tracking (MOT) plays a crucial role in various UAV applications,
ranging from real-time suspect tracking conducted by the police
to interactions between humans and drones 
\cite{lee2018forecasting, mohaimenianpour2018hands, zhao2021research}.
The primary goal of MOT in such applications is to accurately estimate and trace the trajectories of multiple objects
in real-time video streams captured by cameras mounted on drones.
However, MOT in UAV applications poses unique challenges
due to the fast motion of UAVs,
as well as the small size of target objects
caused by the high-altitude and wide-angle views of drones.
These inherent challenges can significantly hinder the overall accuracy of object tracking performance.

To address these challenges,
most contemporary MOT approaches
employ a fusion of motion cues and deep learning-based appearance similarity features.
Nonetheless, the dynamic and unconstrained mobility characteristic of UAVs
introduces notable hurdles for conventional motion cue processing techniques like the Kalman Filter \cite{kalman1960new}.
Additionally, rapid changes in viewpoints and the small scale of objects
can present difficulties for approaches relying on appearance similarity,
including the application of deep learning-based re-identification matching methods.
This is primarily due to the frequent occurrence of low-confidence detections for small-scale objects,
resulting in higher rates of both false positives and false negatives.
Furthermore, modern deep learning-based re-identification modules often
struggle to obtain high-quality visual features in scenarios
where small-scale or partially obscured objects are involved,
primarily due to the restricted image coverage of these objects.
Therefore, this paper outlines two main objectives to tackle these challenges:
(1) Develop a more robust technique to compensate for the irregular motion of UAVs.
(2) Effectively manage the frequent occurrence of low-confidence detections in UAV applications,
primarily stemming from blurred images, small object sizes, and occlusions.

We begin by introducing a simple yet effective technique called ``UAV Motion Compensation''
to address our first objective.
This approach mitigates UAV motion by applying adjustments to the bounding boxes used
for object tracking while preserving their aspect ratios.
It can improve the traditional Intersection over Union (IoU) matching
by leveraging the Kalman Filter to handle erratic UAV motion.

\begin{figure}[t]
\centering
\includegraphics[width=\linewidth]{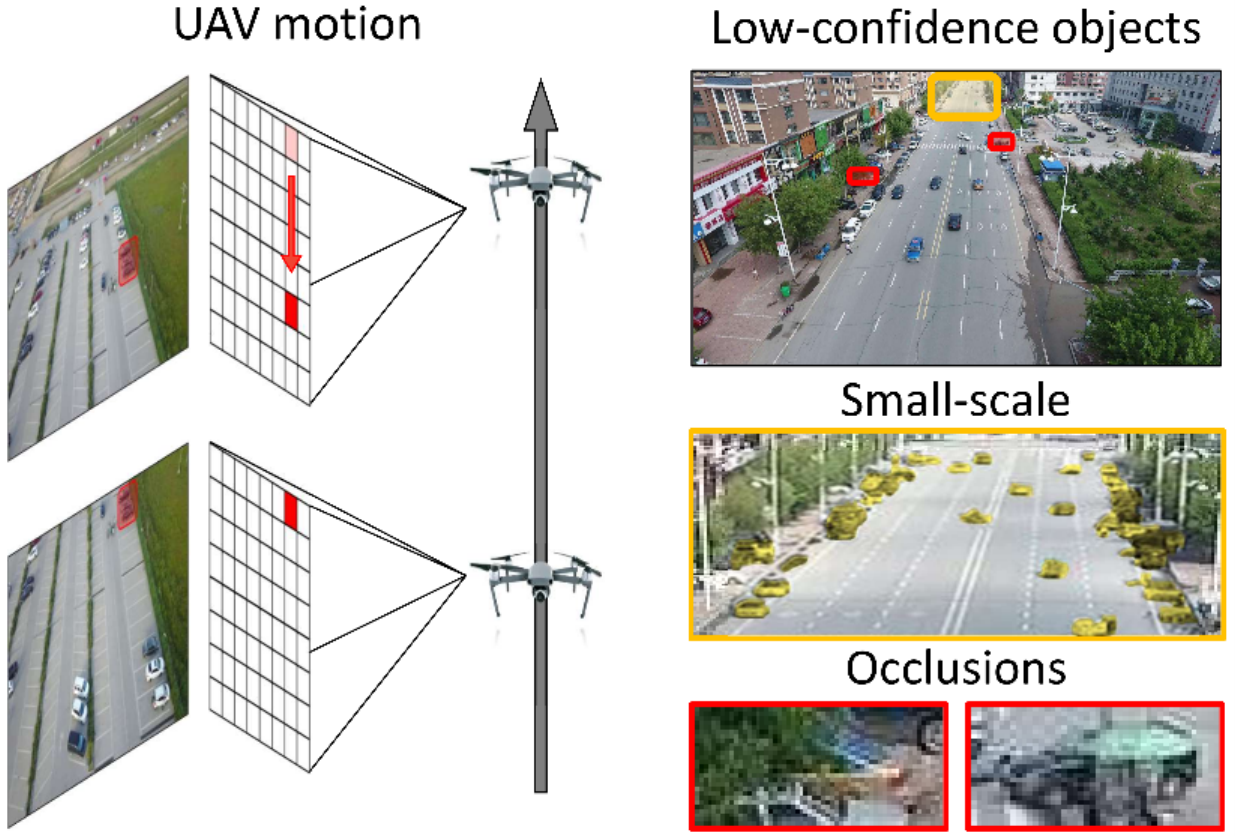}
   \caption{
Illustration of challenges encountered during multi-object tracking in UAVs.
These primarily include irregular patterns of UAV motion,
along with low-confidence detections
due to small scale objects and occlusions.
   }
\label{fig:challenges}
\end{figure}

Concerning our second objective, the most straightforward approach would likely be
to initiate tracking exclusively with high-confidence detections,
a strategy commonly employed in most existing MOT methods.
However, this method is prone to missing the tracking of many target objects,
resulting in an increased number of false negatives.
As a solution, this paper suggests initiating tracking for target objects based on low-confidence detections.
However, this tracking strategy is expected to produce more false positives.
Therefore, there is a need for a method to reduce false positives while effectively tracking a larger number of target objects,
including small-scale objects with low-confidence scores.
To address this issue, we propose employing traditional appearance matching algorithms
based on hand-crafted appearance features,
such as color histograms and scaled image Mean Squared Error (MSE),
for the effective association of these low-confidence detections.
Our hypothesis is that these traditional features can enhance the data association step,
particularly in scenarios where obtaining high-quality deep learning-based visual features is challenging,
for instance, when dealing with small-scale or partially obscured objects.
In contrast to deep learning-based features, traditional hand-crafted features
can be more robust for image matching in situations with high occlusions and small-scale objects.
In addition, the utilization of traditional techniques offers additional benefits,
as they do not require a large volume of training data,
and tend to be very general,
and cost-efficiency in contrast to deep learning-based features \cite{o2020deep}.

To assess the efficacy of our approach,
we conducted experiments to measure object tracking performance on two UAV-specific datasets (VisDrone2019 and UAVDT)
as well as a general dataset (MOT17), achieving state-of-the-art results.
Moreover, throughout our experimentation
we identified annotation errors in the UAVDT dataset, including inaccuracies and omissions.
Consequently, we rectified numerous erroneous annotations and supplemented annotations
for the many instances of missing data in the UAVDT dataset.
We plan to make our enhanced version of the annotations accessible to the public,
thereby facilitating further research in this field.
To summarize, this paper offers the following contributions:
\begin{itemize}
\setlength\itemsep{0em}
    \item We present an approach for UAV multi-object tracking that adeptly handles inherent challenges,
    including rapid and irregular UAV motion, small-scale objects, and occlusions.
    \item We introduce a novel strategy for initiating tracking from low-confidence detections, particularly beneficial in UAV contexts.
    To manage these low-confidence detections,
    we suggest revisiting traditional appearance matching algorithms based on hand-crafted features.
    \item We provide comprehensive evaluation of our approach on multiple datasets, including VisDrone2019, UAVDT, and MOT17, demonstrating its effectiveness and superior performance.
    \item We have identified and corrected missing and incorrect annotations in the UAVDT dataset, contributing to more accurate evaluation and further research.
\end{itemize}

\section{Related Work}

\subsection{Object Detection in Multi-Object Tracking (MOT)}
Recent vision-based multi-object tracking systems mainly use a tracking-by-detection approach,
a two-step process of detection and tracking \cite{zhang2022bytetrack, aharon2022bot, bergmann2019tracking, wang2020towards, zhou2020tracking, zhang2021fairmot}.
This method involves first detecting objects in video frames and then associating these detections over time to track their movement.
Our research focuses on enhancing the association phase of this approach,
specifically tailored to the unique challenges of UAV footage.

\subsection{Data association for Multi-Object Tracking}
The data association phase matches detection results with tracklets.
It typically uses motion and appearance cues to compute similarities
between objects and applying a matching strategy.
\textbf{Motion cue}
Motion-based strategies utilize object or camera movement information (motion cue) to effectively track objects.
SORT \cite{bewley2016simple} stands as a pioneering work in this field,
employing Kalman Filters to predict object locations and compare these predictions with new detections.
Tracktor \cite{bergmann2019tracking} suggested camera motion compensation
by aligning frames through image registration \cite{evangelidis2008parametric}.
OC-SORT \cite{cao2023observation} enhances the traditional Kalman Filter approach by focusing on object-centric adjustments, yielding improved tracking performance in scenarios with irregular motion.
BoTSORT \cite{aharon2022bot}
proposed correcting Kalman Filter's predicted bounding box locations
using camera motion compensation based on feature point tracking \cite{shi1994good, bouguet2001pyramidal}.
However, prior studies often fall short in handling extensive motion in UAV footage.
This study tackles this issue by introducing a straightforward yet efficient UAV-specific motion compensation technique to enhance tracking accuracy and mitigate bounding box distortion.

\textbf{Appearance cue}
Relying solely on motion cues can be challenging
especially when bounding box predictions are imprecise.
Consequently, strategies incorporating appearance cues have emerged to address these challenges.
Earlier methods used simple techniques like color histograms \cite{ju2010mean}, but modern approaches like DeepSORT \cite{wojke2017simple} and JDE \cite{wang2020towards} use advanced neural networks to analyze object features. 
Similarly, MOTDT \cite{chen2018real} used a Re-ID network \cite{zhao2017deeply} for feature extraction,
while Tracktor \cite{bergmann2019tracking} utilized a siamese network.
More recently, several attempts have been made to simultaneously train object detectors
and Re-ID modules to enhance feature learning,
as evidenced by JDE \cite{wang2020towards} and FairMOT \cite{zhang2021fairmot}.
Such approaches, however, have failed to effectively address the unique challenges presented by UAV-captured videos,
including issues with small and fast objects as well as the common occurrence of occluded objects. 
Consequently, they often encounter difficulties in accurately tracking target objects.
Hence, in this paper, we propose revisiting traditional hand-crafted appearance features
like color histograms and scaled image MSE.
Through experimentation, we demonstrate that these traditional appearance cues can exhibit enhanced robustness,
particularly when dealing with small-scale or partially obscured objects.

\textbf{Matching strategy}
In terms of the matching strategy, the Hungarian algorithm,
originally introduced by Kuhn in 1955 \cite{kuhn1955hungarian},
is considered the standard-de-facto approach for associating detections with tracklets.
DeepSORT \cite{wojke2017simple} took this a step further
by introducing a cascaded matching algorithm that gives priority to recent tracklets.
MOTDT \cite{chen2018real} first applies the appearance cue, and then employs IoU for any tracklets that remain unmatched.
ByteTrack \cite{zhang2022bytetrack} introduces a secondary algorithm
specifically designed to associate detections with lower confidence to existing tracklets.
More recently, Transformer based architectures \cite{meinhardt2022trackformer, zeng2022motr} employ track queries to find object locations, enabling implicit matching through attention mechanisms without relying on the Hungarian algorithm.
However, so far, very little attention has been paid to
effectively handling and utilizing objects detected with low confidence scores, a common occurrence in UAV application scenarios.
This is likely because using low-confidence detections without careful consideration of a new matching strategy
may lead to more false positives.
In contrast to prior studies, this paper proposes initializing new tracks from low-confidence detections
by effectively managing them through traditional handcrafted feature-based appearance matching.

\subsection{Multi-object Tracking on UAVs}
For MOT in UAV videos,
unique challenges arise from small object sizes and blurred images.
Therefore, in recent years, competitions like VisDrone have emerged \cite{fan2020visdrone},
driving the development of methods tailored for UAV-specific scenarios.
V-IOU Tracker \cite{bochinski2018extending} enhances tracking using visual cues in the absence of detections.
HMTT \cite{pan2019multi} employs a hierarchical approach with a re-identification subnet,
while GIAOTracker \cite{du2021giaotracker} introduces the NSA Kalman Filter, which adjusts noise scale during state updates and several post-processing approaches.
In contrast, UAVMOT \cite{liu2022multi} and FOLT \cite{yao2023folt} provide online tracking solutions.
UAVMOT employs an adaptive motion filter for challenging scenarios, and FOLT utilizes optical flow to track small objects.
In contrast to these methods primarily refining motion tracking for UAV scenarios,
our framework integrates improvements in both motion and appearance cues, along with matching strategies,
to create a more accurate and robust tracking system for the challenging environment of UAV footage.

\begin{algorithm}[!ht]
\SetAlgoLined

\SetKwInOut{Input}{Input}
\SetKwInOut{Output}{Output}

\Input{A video sequence $V$; object detector $Det$; detection score threshold $\tau$; tracking initiate threshold $\rho$}
\Output{Tracks $\Tau$ of the video}

\BlankLine
Initialization: $\Tau$ $\leftarrow \emptyset$ \\
\For{frame $f_k$ in V}{
    $D_k$ $\leftarrow$ Det($f_k$) \\
    $D_{high}$ $\leftarrow \emptyset$ \\
    $D_{low}$ $\leftarrow \emptyset$ \\
    \BlankLine

    \For{$d$ in $D_k$}{
        \eIf{$d.det\_score > \tau$}{
            $D_{high}$ $\leftarrow$ $D_{high}$ $\cup$ \{$d$\}
        }
        {
            $D_{low}$ $\leftarrow$ $D_{low}$ $\cup$ \{$d$\}
        }
    }
    \BlankLine

    \begingroup
    \color{brown}
    /\# UAV motion compensation \textbf{Section 3.A} \#/\\
    /\# Figure 2 \#/\\
    $M_k$ $\leftarrow$ UAV motion from $f_{k-1}$ to $f_k$ \\
    \For{$t$ in $\Tau$}
    {
        $t$ $\leftarrow$ KalmanFilter($t$) \\
        $t$ $\leftarrow$ $M_k$($t$) \\
    }
    \endgroup
    \BlankLine
 
    /\# 1st association: High detections \#/ \\
    Associate $\Tau$ and $D_{high}$ \\
    $D_{high\_remain}$ $\leftarrow$ unmatched $D_{high}$ \\
    $T_{remain}$ $\leftarrow$ unmatched $\Tau$ \\
    \BlankLine

    \begingroup
    \color{teal}
    /\# 2nd association: Low detections \textbf{Section 3.B} \#/\\
    /\# Figure 3 \#/\\
    Associate $\Tau_{remain}$ and $D_{low}$\\
    $D_{low\_remain}$ $\leftarrow$ unmatched $D_{low}$\\
    $\Tau_{re\_remain}$ $\leftarrow$ remaining tracks from $\Tau_{remain}$\\
    \endgroup
    \BlankLine

    /\# Delete unmatched tracks \#/\\
    $\Tau$ $\leftarrow$ $\Tau$ $\setminus$ $\Tau_{re\_remain}$ \\
    \BlankLine
    
    \begingroup
    \color{purple}
    /\# Initialize new tracks \textbf{Section 3.C} \#/\\
    /\# Init $\Tau$ from unmatched high-score detections \#/\\
    \For{$d$ in $D_{high\_remain}$}{
        $\Tau$ $\leftarrow$ $\Tau$ $\cup$ \{$d$\}
    }
    \BlankLine
    
    /\# Init $\Tau$ from low-score detections \#/\\
    \For{$d$ in $D_{low\_remain}$}{
        \If{$d$.appearance similarity with $D_{high}$ $>$ $\rho$}{
            $\Tau$ $\leftarrow$ $\Tau$ $\cup$ \{$d$\}
        }
    }
    \endgroup
    }

\Return{$\Tau$}
\caption{Pseudo-code of SFTrack.}
\label{alg:tracking algorithm}
\end{algorithm}

\section{Approach}
In this section, we introduce our tracking strategy \textbf{SFTrack} (\textbf{S}mall and \textbf{F}ast-moving Objects \textbf{Track}ing),
designed to address the frequent occurrence of low-confidence detections in UAV applications.
Unlike previous methods, such as \cite{aharon2022bot, zhang2022bytetrack, wojke2017simple, bewley2016simple, chen2018real},
which take a conservative approach by starting tracking only on high-confidence detections,
we actively leverage low-confidence detections as the starting point for tracking.
Furthermore, we enhance tracking accuracy
by revisiting traditional appearance matching algorithms for data association,
such as color histogram similarity and scaled image MSE.
Additionally, we employ a motion compensation technique
designed for irregular and rapid movements in UAV footage.

The method details are outlined following the steps in Alg. \ref{alg:tracking algorithm}.
Our tracking algorithm takes a video sequence $V$ as input,
along with an object detector $Det$,
with defined parameters:
a detection confidence threshold $\tau$,
and a low-confidence detection tracking threshold $\rho$.
The output comprises a set of tracks $\Tau$
each containing bounding box coordinates across frames
while preserving the identity of the objects.

This process begins by employing the object detector
to predict bounding boxes and confidence scores.
Subsequently, similar to BYTE \cite{zhang2022bytetrack},
we then categorize detections into high-confidence $D_{high}$ and low-confidence $D_{low}$ groups,
guided by the score threshold $\tau$ (refer to lines 1-12 in Alg. \ref{alg:tracking algorithm}).

\begin{figure}[!t]
\centering
\includegraphics[width=\linewidth]{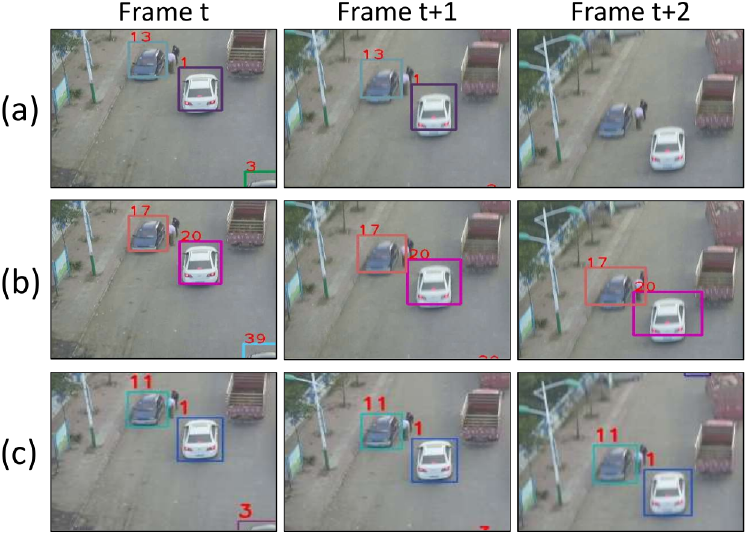}
   \caption{
    Comparison of tracking results on a low-altitude video.
    (a) ByteTrack without motion-compensation (MC), fails to maintain tracks.
    (b) BoTSORT's MC leads to distorted bounding boxes.
    (c) SFTrack with UAV MC, ensures consistent and accurate tracking.
   }
\label{fig:uav motion}
\end{figure}

\subsection{UAV Motion Compensation}
After separating low-score and high-score detection boxes,
we apply an enhanced motion compensation algorithm by incorporating a simple yet effective idea,
drawing inspiration from Aharon et al.'s method \cite{aharon2022bot}.
While the conventional motion compensation is effective,
it often falters in low-altitude UAV scenes
due to errors predominantly caused by bounding box ratio distortion.
Our solution is a straightforward adjustment, that preserving the bounding box ratio.
To implement this, we calculate the affine transformation matrix $M_k$ capturing camera motion by extracting feature points \cite{shi1994good}
and utilizing sparse optical flow \cite{bouguet2001pyramidal} to track these features.
This matrix, encompassing scale, rotation, and translation components,
guides the transformation of bounding boxes.
Subsequently, we update the state vector of tracks $\Tau$ using Kalman Filter predictions,
applying the $M_k$ to this updated state vector.
Furthermore, to prevent significant distortion in the bounding box ratio during this process,
we enforce constraints on the scale factor in $M_k$,
specifically selecting the larger scale factor
between the x and y axes and uniformly applying it to both axes.
This approach maintains a consistent bounding box ratio across frames,
even during intense UAV camera motion at low-altitudes (Figure~\ref{fig:uav motion}).


After compensating for UAV motions,
we first associate all tracks $\Tau$ with high-confidence detections $D_{high}$
using IoU and cosine similarity of Re-ID features \cite{luo2019bag}.
The Hungarian algorithm \cite{kuhn1955hungarian} is employed for the association
based on the product of these measures.
Unmatched high-confidence detections then form set $D_{high\_remain}$,
and the remaining tracks constitute set $T_{remain}$
(see lines 20-23 in Alg. \ref{alg:tracking algorithm}).

\subsection{Matching low-confidence detections}
Following the initial association,
a second association is performed between
the remaining tracks $T_{remain}$ and low-confidence detections $D_{low}$.
In contrast to the first association, which relied on deep learning-based Re-ID features,
in this phase, we propose to utilize traditional hand-crafted appearance features
for these low-confidence detections (see lines 24-27 in Alg. \ref{alg:tracking algorithm}).
Specifically, we employ color histogram similarity and scaled image MSE as appearance cue.
Therefore, associations are established through the Hungarian algorithm,
relying on the product of three key values:
IoU, color histogram similarity, and scaled image MSE.

Color histograms offer an efficient mechanism
to capture the color distribution characteristics
of an object within an image.
We categorize the color intensity into 8 uniformly spaced levels (0-31, 32-63, ..., 224-255)
for each color channel (R, G, B),
and record the frequency of pixel color values.
The color histogram similarity is then calculated by the Bhattacharyya distance \cite{kailath1967divergence}.
Additionally, we employ scaled image MSE to measure image dissimilarity. 
To ensure a fair comparison regardless of object scale,
images cropped by bounding boxes are first resized to the same size.
Subsequently, the MSE is computed between these two images.
To convert MSE to a similarity score within a range of 0-1, we subtract the normalized MSE value from 1.

\begin{figure}[!t]
\includegraphics[width=\linewidth]{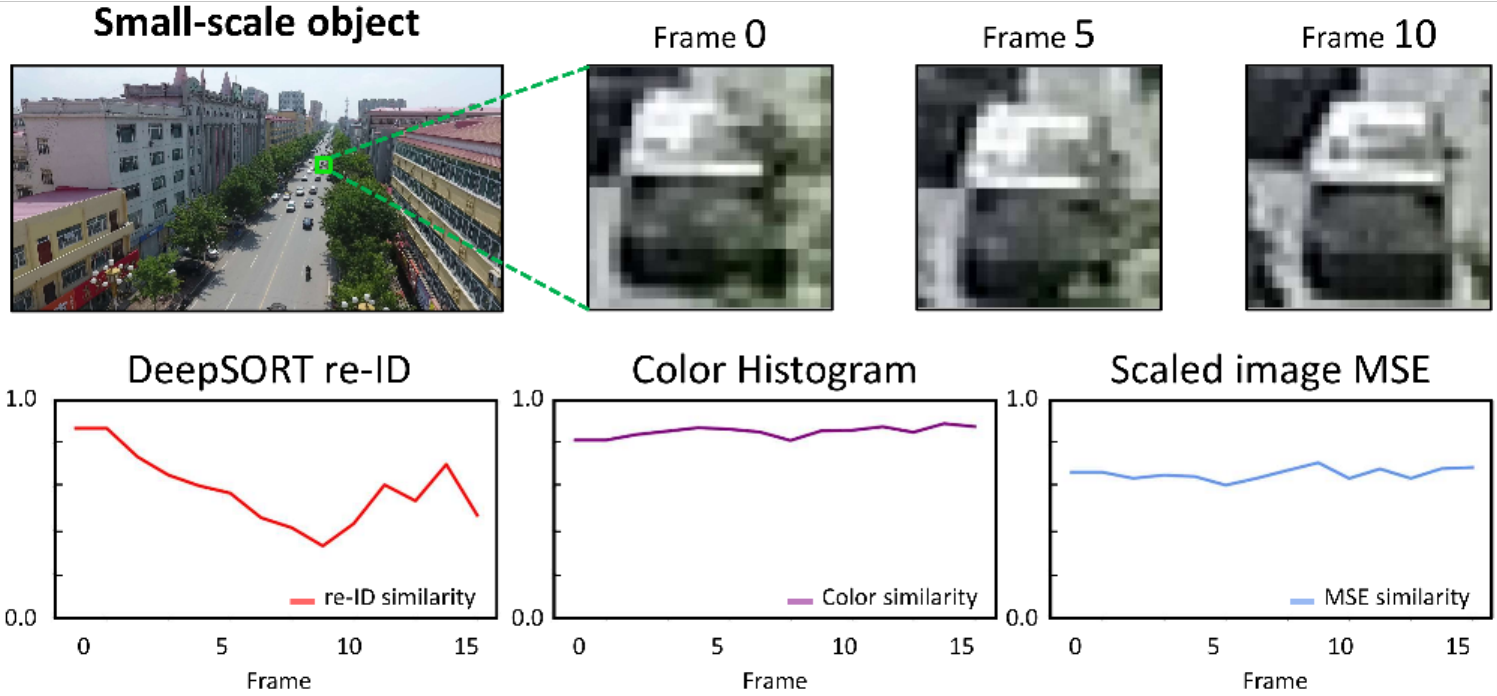}
   \caption{
   This figure compares appearance similarity methodologies for small-scale objects.
   Re-ID faces challenges in low-resolution and occlusion scenarios,
   while color histogram and MSE matching offer consistent similarity throughout the frame.
   }
\label{fig:small scale object}
\end{figure}

The motivation behind this is rooted in the limitations of deep learning methods,
including Re-ID, which frequently encounter challenges in low-resolution settings or when objects are occluded.
This is due to their inclination to prioritize the foreground object, 
potentially overlooking the object of interest when it is located in the background.
In contrast, traditional matching algorithms, like color histogram similarity and scaled image MSE,
consider the complete image, regardless of occlusion.
Thus, these methods show superior performance in occlusion and low-resolution scenarios
due to their holistic use of image information.
As illustrated in Figure~\ref{fig:small scale object},
our approach to low-confidence detection association outperforms deep learning-based methods,
particularly in low-resolution and occlusion scenarios.

After this phase of matching low-confidence detections,
any tracks that remain unmatched following both the high- and low-confidence detection matching phases
are removed from the tracking pool.
(see Alg. \ref{alg:tracking algorithm}, lines 28-29).
In this context, we provide a grace period before removing these tracks,
waiting for 30 frames where they don't find a match in a row.
This waiting period, influenced by the BYTE approach \cite{zhang2022bytetrack},
prevents premature termination of a valid track
due to temporary occlusions or brief detection failures.

\subsection{Initiation of tracking from low-confidence detections}
Finally, in contrast to previous approaches such as BYTE \cite{zhang2022bytetrack},
we introduce a distinctive method
by initiating new tracks not only from unmatched high-confidence detections ($D_{high\_remain}$)
but also from low-confidence detections (see Alg. \ref{alg:tracking algorithm}, lines 30-39).
However, initiating tracks from low-confidence detections poses a challenge
as this new strategy is expected
to generate more redundant, unnecessary, and potentially intrusive object candidates.

To resolve this challenge, we evaluate low-confidence detections
by comparing their appearance similarity with high-confidence detections from the same object category using Re-ID features.
This process functions as a filtering mechanism,
allowing us to selectively initiate new tracks exclusively from unmatched low-confidence detections ($D_{low\_remain}$)
that surpass a predetermined appearance similarity threshold ($\rho$) with high-confidence detections.
This double-checking procedure enhances the reliability of low-confidence detections,
improving tracking of small-scale objects
and enhancing the overall performance of our system.

\section{Datasets and Evaluation Metrics}
In this study, we conducted experiments on three distinct datasets:
two UAV-specific datasets, VisDrone2019 and UAVDT, and a general dataset, MOT17.
For evaluation, we employed CLEAR metrics \cite{bernardin2008evaluating},
including Multiple Object Tracking Accuracy (MOTA), False Positives (FP), False Negatives (FN), and ID switches (IDs).

\subsection{Datasets}
\noindent \textbf{VisDrone2019}\cite{zhu2021detection} is employed for five tasks:
Object Detection in Images, Object Detection in Videos, Single-Object Tracking, Multi-Object Tracking (MOT),
and Crowd Counting in videos captured by UAVs.
Among them, this paper specifically concentrates on the MOT task within the VisDrone2019 dataset,
comprising 56 sequences in the training set,
7 sequences in the validation set, and 17 sequences in the test-dev set.
To ensure a fair comparison,
we limit our focus to five object categories:
pedestrian, car, van, truck, and bus, aligning with the object categories used in \cite{liu2022multi}.

\noindent \textbf{UAVDT}\cite{du2018unmanned},
is a UAV Detection and Tracking benchmark,
offering 30 training sequences and 20 testing sequences
exclusively focusing on car tracking for the MOT task.
Unlike other datasets, UAVDT provides annotations with environmental attributes such as weather, altitude, and camera view,
captured under diverse conditions.
%

\noindent \textbf{MOT17}\cite{milan2016mot16}
serves as a prominent dataset in the field of multi-object tracking.
It covers a wide array of urban scenarios involving pedestrians, vehicles, and other dynamic entities.
We adhere to the protocol specified in \cite{zhou2020tracking, zhang2022bytetrack},
splitting the training set equally for training and validation,
and focusing on the pedestrian category for the MOT task.
This dataset is the only non-UAV dataset used in our setup.

\begin{figure}[!t]
\centering
\includegraphics[width=\linewidth]{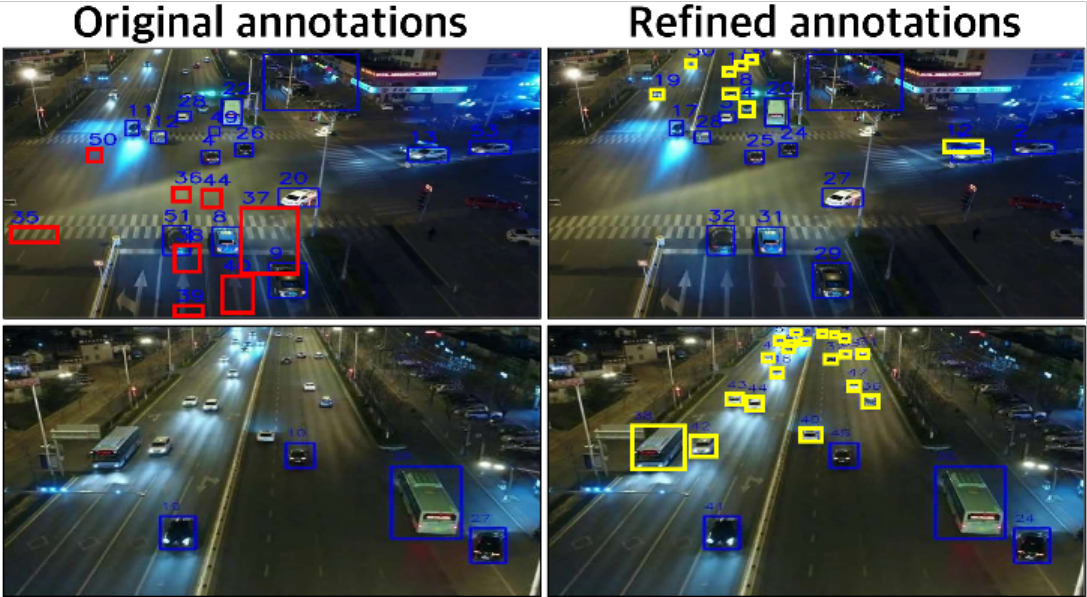}
   \caption{
This image provides a comparative visualization of the Original UAVDT and our Refined UAVDT annotations.
Errors in the original annotations, which do not correspond to actual objects, are highlighted in `red'.
The `yellow' markers represent our additional annotations for visible objects within the Refined UAVDT dataset.
   }
\label{fig:UAVDT annotations}
\end{figure}

\subsection{UAVDT refinement}
During our study,
we identified significant annotation errors in the UAVDT dataset,
as illustrated in Figure~\ref{fig:UAVDT annotations}.
Significantly, certain objects lacked annotations,
while numerous annotations either commenced before the object appeared or persisted beyond its exit in the image frame.
To rectify these issues and improve the accuracy of our study, we refined annotations within 4721 frames. 
This effort, reflected in the `Refined UAVDT,' introduced 43,981 additional annotations,
elevating the count from 340,906 to 384,887 objects, along with 55 extra tracks.
We plan to publicly release this enhanced dataset for future research.

\subsection{Implementation details}
Our experiments are conducted on a desktop server
equipped with an Intel Core i9-10900X CPU @ 3.70GHz processor and single NVIDIA GeForce RTX 3090 GPU.
We employed the YOLOX \cite{ge2021yolox} object detector, pre-trained on the COCO dataset \cite{lin2014microsoft}.
For VisDrone2019 and UAVDT experiments, the training was conducted on their respective training sets,
with an input video resolution of 1920 $\times$ 1080, adhering to the protocol set by \cite{liu2022multi}.
For the MOT17 experiment, we trained on both the CrowdHuman \cite{shao2018crowdhuman} dataset
and the first half of the MOT17 training set,
while evaluation was performed on the second half of the MOT17 validation set.
The input video resolution for this experiment was 1440 $\times$ 800, in line with \cite{zhang2022bytetrack}.

\subsection{Evaluation metrics}
In this section, we describe how we compute MOTA and IDF1,
kindly providing a reminder.
For further details on CLEAR metrics,
please refer to \cite{bernardin2008evaluating}.
MOTA is a composite metric that considers FP, FN, and IDs and is calculated as follows:
\begin{equation}
\footnotesize
MOTA = 1 - \frac{(FP + FN + IDs)}{GT} \times 100 %
\label{eq:MOTA}
\end{equation}
Here, GT denotes numbers of ground-truth objects.

We also adopt the IDF1 score from \cite{ristani2016performance},
which primarily assesses the identity preservation capability, recognizing that in the MOTA score,
the impact of larger quantities of FP and FN is often more significant than that of ID
\begin{equation}
\footnotesize
IDF1 = \frac{2 \cdot IDTP}{GT + C}
\label{eq:IDF1} 
\end{equation}
Here, IDTP represents the number of True Positives with correct identities, and C refers to the total computed detections.
Additionally, we applied the standard metrics, Mostly Tracked (MT) and Mostly Lost (ML) \cite{li2009learning}.
MT and ML represent the proportions of ground truth trajectories
that are covered by computed trajectories with correct and incorrect identities for at least 80\% of their lifespan, respectively.

\begin{table}[!t]
\caption{
Comparison of data association methods on \textbf{VisDrone2019} test-dev set.
(*indicates methods using their own detector).}
\label{table:VisDrone2019-test-results}
\begin{adjustbox}{max width=\columnwidth}
\begin{tabular}{ l | c  c  c  c  c  c  c }
\toprule

\rule{0pt}{0.3cm} Method & MOTA$\uparrow$ & IDF1$\uparrow$ & IDs$\downarrow$ & FP$\downarrow$ & FN$\downarrow$ & MT$\uparrow$ & ML$\downarrow$ \\
\hline
\rule{0pt}{0.3cm} SORT \cite{bewley2016simple}     & 25.7 & 46.4 & 865 & \textbf{9592} & 162216 & 366 & 856  \\
\rule{0pt}{0.3cm} MOTDT \cite{chen2018real}   & 34.9 & 44.8 & 1692 & 13280 & 136283 & 440 & 805  \\
\rule{0pt}{0.3cm} DeepSORT \cite{wojke2017simple} & 35.1 & 47.0 & 934 & 12773 & 137078 & 421 & 810  \\
\rule{0pt}{0.3cm} UAVMOT* \cite{liu2022multi} & 36.1 & 51.0 & 2775 & 27983 & 115925 & 520 & 574  \\
\rule{0pt}{0.3cm} BoTSORT \cite{aharon2022bot} & 39.6 & 53.6 & 1066 & 21333 & 118003 & 581 & 659  \\
\rule{0pt}{0.3cm} OCSORT \cite{cao2023observation} & 41.3 & 52.4 & 941 & 12985 & 122445 & 429 & 703 \\
\rule{0pt}{0.3cm} FOLT* \cite{yao2023folt} & 42.1 & 56.9 & 800 & 24105 & 107630 & - & -  \\
\rule{0pt}{0.3cm} ByteTrack \cite{zhang2022bytetrack} & 42.3 & 53.4 & 989 & 15689 & 117539 & 586 & 675  \\
\hline
\rule{0pt}{0.3cm} \textbf{SFTrack} (Ours)    & \textbf{47.2} & \textbf{62.1} & \textbf{557} & 27159 & \textbf{94910} & \textbf{753} & \textbf{518} \\

\bottomrule
\end{tabular}
\end{adjustbox}
\end{table}

\begin{table}[!t]
\caption{Comparison of data association methods on \textbf{UAVDT} test set.
(*indicates methods using their own detector).}
\label{table:UAVDT-test-results}
\begin{adjustbox}{max width=\columnwidth}
\begin{tabular}{ l | c  c  c  c  c  c  c }
\toprule

\rule{0pt}{0.3cm} Method & MOTA$\uparrow$ & IDF1$\uparrow$ & IDs$\downarrow$ & FP$\downarrow$ & FN$\downarrow$ & MT$\uparrow$ & ML$\downarrow$ \\
\hline
\rule{0pt}{0.3cm} SORT \cite{bewley2016simple}     & 32.4 & 55.1 & 165 & 45633 & 179092 & 442 & 405  \\
\rule{0pt}{0.3cm} MOTDT \cite{chen2018real}   & 32.5 & 56.0 & 422 & 55664 & 168754 & 487 & 363  \\
\rule{0pt}{0.3cm} DeepSORT \cite{wojke2017simple} & 33.1 & 56.4 & 311 & 54016 & 168357 & 480 & 365  \\
\rule{0pt}{0.3cm} UAVMOT* \cite{liu2022multi} & 46.4 & 67.3 & 456 & 66352 & 115940 & 624 & 221  \\
\rule{0pt}{0.3cm} OCSORT \cite{cao2023observation} & 47.0 & 67.4 & \textbf{148} & 56309 & 119892 & 559 & 156 \\
\rule{0pt}{0.3cm} BoTSORT \cite{aharon2022bot} & 47.3 & 68.8 & 417 & 75547 & 99365 & 667 & 174  \\
\rule{0pt}{0.3cm} ByteTrack \cite{zhang2022bytetrack} & 47.5 & 69.4 & 159 & 76181 & 98415 & 688 & 167  \\
\rule{0pt}{0.3cm} FOLT* \cite{yao2023folt} & 48.5 & 68.3 & 338 & \textbf{36429} & 155696 & - & -  \\
\hline
\rule{0pt}{0.3cm} \textbf{SFTrack} (Ours)     & \textbf{49.5} & \textbf{70.3} & 209 & 72866 & \textbf{95935} & \textbf{709} & \textbf{152} \\

\bottomrule
\end{tabular}
\end{adjustbox}
\end{table}

\section{Experiments and Results}

\subsection{Comparisons with other state-of-art approaches}
We compared our approach with other prominent association methods using three datasets: VisDrone2019, UAVDT, and MOT17.
However, it should be acknowledged that UAVMOT \cite{liu2022multi} and FOLT \cite{yao2023folt}
do not strictly fall under the category of data association methods,
thereby limiting their evaluation in an identical environment.
Nevertheless, we referenced evaluation protocols from their original papers,
and strive to align our evaluation settings, considering factors such as categories and input image resolution.

\noindent \textbf{UAV Datasets}
On the UAV datasets, our SFTrack outperforms others significantly across most evaluation metrics.
For example, our method achieved a 4.9\% improvement in MOTA over the top-performing ByteTrack and a 5.2\% increase in IDF1 score compared to FOLT, the leading IDF1 scorer on VisDrone2019.
However, we acknowledge the trade-offs of our approach, specifically between a reduction in false negatives (FN) and an increase in false positives (FP).
This trade-off stems from our methodology,
which actively utilizes low-confidence detections and employs traditional appearance matching algorithms for association.
However, it is worth noting that the proposed SFTrack demonstrates
more balanced tracking performance
compared to previous state-of-the-art methods.
Moreover, in many tracking scenarios, it is more crucial not to miss target objects than to briefly track additional objects,
including those that are not of interest.

\begin{table}[!t]
\caption{Results on \textbf{Refined UAVDT} test set using "same detector".}
\label{table:Refined-UAVDT-test-results}
\begin{adjustbox}{max width=\columnwidth}
\begin{tabular}{ l | c  c  c  c  c  c  c }
\toprule

\rule{0pt}{0.3cm} Method & MOTA$\uparrow$ & IDF1$\uparrow$ & IDs$\downarrow$ & FP$\downarrow$ & FN$\downarrow$ & MT$\uparrow$ & ML$\downarrow$ \\
\hline
\rule{0pt}{0.3cm} SORT \cite{bewley2016simple}     & 35.6 & 55.3 & 217 & \textbf{33073} & 207951 & 419 & 448  \\
\rule{0pt}{0.3cm} MOTDT \cite{chen2018real}   & 36.4 & 56.4 & 590 & 41505 & 196014 & 464 & 402  \\
\rule{0pt}{0.3cm} DeepSORT \cite{wojke2017simple} & 36.8 & 56.7 & 418 & 40166 & 195926 & 454 & 405  \\
\rule{0pt}{0.3cm} OCSORT \cite{cao2023observation} & 52.1 & 69.8 & \textbf{186} & 37908 & 141340 & 535 & 192 \\
\rule{0pt}{0.3cm} BoTSORT \cite{aharon2022bot} & 52.5 & 69.7 & 441 & 56067 & 121304 & 646 & 201  \\
\rule{0pt}{0.3cm} ByteTrack \cite{zhang2022bytetrack} & 53.0 & 70.3 & 215 & 56021 & 119674 & 668 & 193  \\
\hline
\rule{0pt}{0.3cm} \textbf{SFTrack} (Ours)     & \textbf{55.3} & \textbf{71.3} & 284 & 49946 & \textbf{117224} & \textbf{682} & \textbf{184} \\

\bottomrule
\end{tabular}
\end{adjustbox}
\end{table}

\begin{table}[!t]
\caption{Results on \textbf{MOT17} validation set using "same detector".}
\label{MOT17 val results}
\begin{adjustbox}{max width=\columnwidth}
\begin{tabular}{ l | l | c  c  c  c  c  }
\toprule

\rule{0pt}{0.3cm} Detector & Method & MOTA$\uparrow$ & IDF1$\uparrow$ & IDs$\downarrow$ & FP$\downarrow$ & FN$\downarrow$  \\
\hline
\rule{0pt}{0.3cm} \multirow{8}{*}{\parbox{2cm}{YOLOX-x \\ (AP = 0.624)}} &SORT     &74.6 & 77.3 & 287 & 2797 & 10592  \\
\rule{0pt}{0.3cm} &DeepSORT   &75.4 & 77.2 & 239 & 3399 & 9641  \\
\rule{0pt}{0.3cm} &MOTDT &75.8 & 77.7 & 263 & \textbf{2689} & 10069  \\
\rule{0pt}{0.3cm} &OCSORT  & 76.0 & 79.5 & 159 & 3325 & 9406 \\
\rule{0pt}{0.3cm} &BoTSORT  & 76.3 & 79.5 & 162 & 3521 & 9088 \\
\rule{0pt}{0.3cm} &ByteTrack  & 76.4 & 79.3 & 164 & 3418 & 9126 \\
\rule{0pt}{0.3cm} &\textbf{SFTrack} (Ours)  &\textbf{76.9} & \textbf{80.1} & \textbf{153} & 3237 & \textbf{9044} \\
\hline
\rule{0pt}{0.3cm} \multirow{8}{*}{\parbox{2cm}{YOLOX-Nano \\ (AP = 0.565)}} &SORT      &69.4 & 67.0 & 705 & 2701 & 13092 \\
\rule{0pt}{0.3cm} &MOTDT & 70.4 & 72.1 & 300 & \textbf{1730} & 13905 \\
\rule{0pt}{0.3cm} &DeepSORT & 71.0 & 73.1 & 311 & 3141 & 12175 \\
\rule{0pt}{0.3cm} &OCSORT & 71.5 & 72.0 & 231 & 2603 & 12486 \\
\rule{0pt}{0.3cm} &BoTSORT & 71.8 & 71.7 & 236 & 2848 & 12158 \\
\rule{0pt}{0.3cm} &ByteTrack & 72.1 & 71.3 & 239 & 2708 & 12066 \\
\rule{0pt}{0.3cm} &\textbf{SFTrack} (Ours) &\textbf{73.0} & \textbf{74.7} & \textbf{220} & 2436 & \textbf{11918} \\


\bottomrule
\end{tabular}
\end{adjustbox}
\end{table}

\begin{figure}[!t]
\centering
\includegraphics[width=0.9\linewidth]{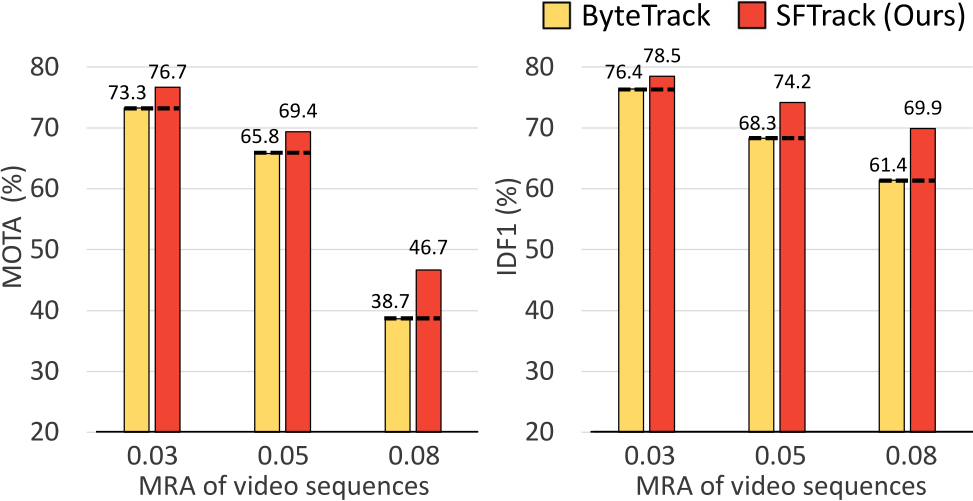}
   \caption{
    Experimental comparision with ByteTrack on VisDrone2019 video sequences.
    Improvements in MOTA and IDF1 scores correlate with higher MRA, indicating the presence of small and fast objects.
   }
\label{fig:MRA Visdrone}
\end{figure}

\begin{figure}[!t]
\centering
\includegraphics[width=0.9\linewidth]{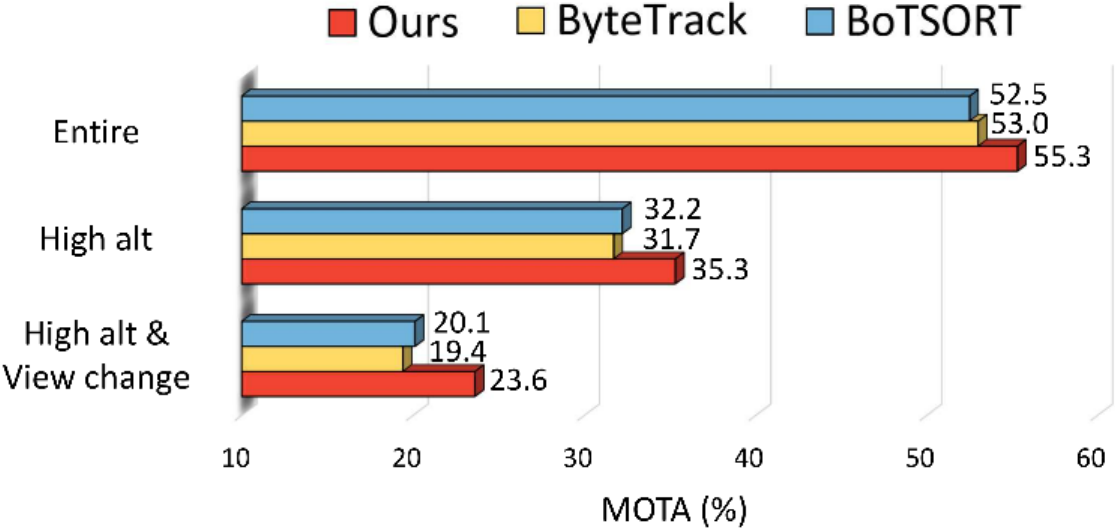}
   \caption{
Experiments on \textbf{Refined UAVDT} under challenging conditions.
   }
\label{fig:UAVDT high-alt}
\end{figure}

\begin{figure}[!t]
\centering
\includegraphics[width=\linewidth]{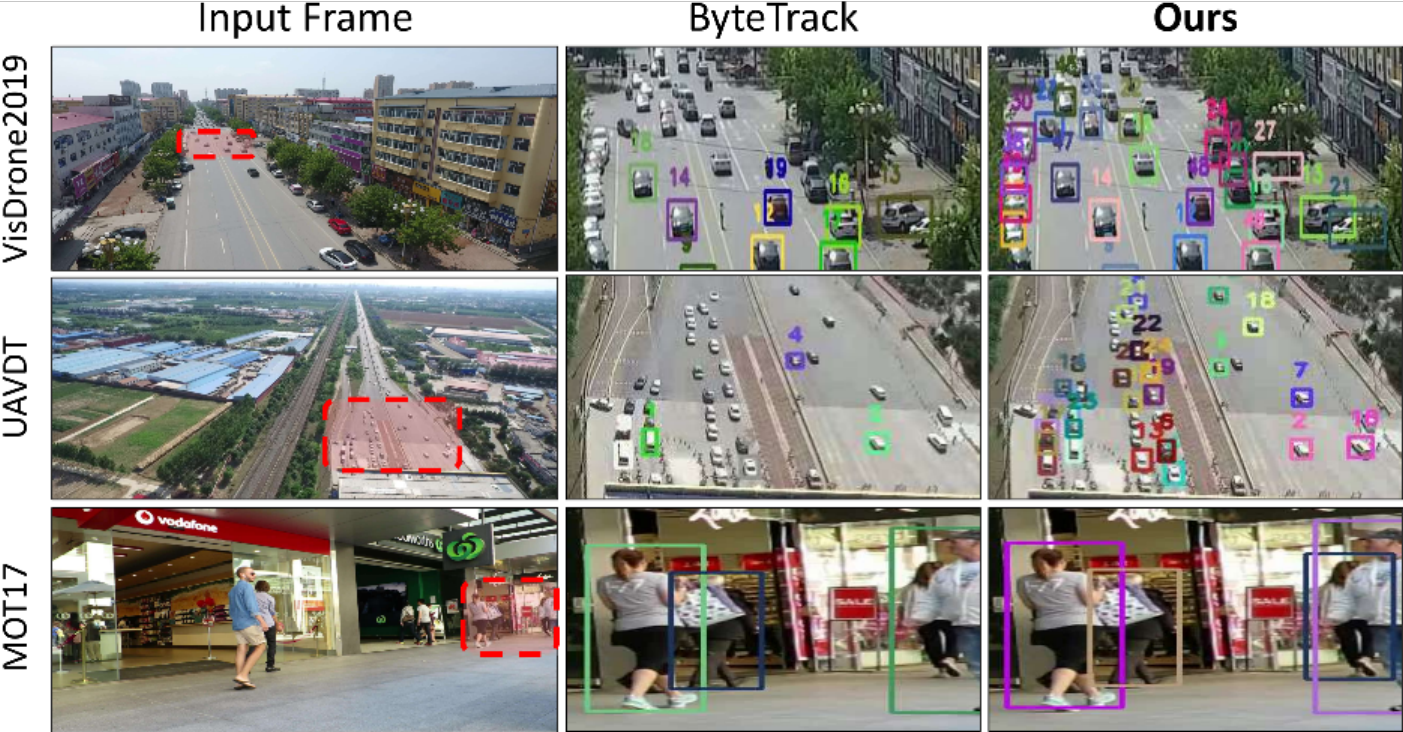}
   \caption{
Qualitative results on VisDrone2019 test-dev, UAVDT test, and MOT17 validation datasets.
Our method significantly outperforms ByteTrack,
especially for occluded or smaller-scale objects.
   }
\label{fig:qualitative result}
\end{figure}

\noindent \textbf{Evaluation on Challenging UAV Subsets}
To explicitly demonstrate SFTrack's enhancement in tracking small and fast-moving objects, we further examined its performance under challenging conditions within the VisDrone2019 and Refined UAVDT datasets.
On VisDrone2019, we specifically selected videos with a high prevalence of small, fast-moving objects, categorized by high Mean Relative Acceleration (MRA) values \cite{yao2023folt}. Our findings, illustrated in Figure~\ref{fig:MRA Visdrone}, reveal that SFTrack's performance substantially improves in scenarios with elevated MRA, underscoring its adeptness at handling rapid object motions and small sizes of objects.
On Refined UAVDT we conducted experiments using videos from high-altitude UAV flights (over 70m from the ground)
and scenarios involving view changes (Figure~\ref{fig:UAVDT high-alt}).
In these challenging conditions, our approach outperformed the next best method by a wider margin,
demonstrating the robustness and efficacy of our methodology.
These experiments affirm SFTrack's capability to significantly improve tracking accuracy for small and fast-moving objects.
Figure~\ref{fig:qualitative result} provides qualitative results supporting this conclusion.

\begin{figure}[!t]
\centering
\includegraphics[width=\linewidth]{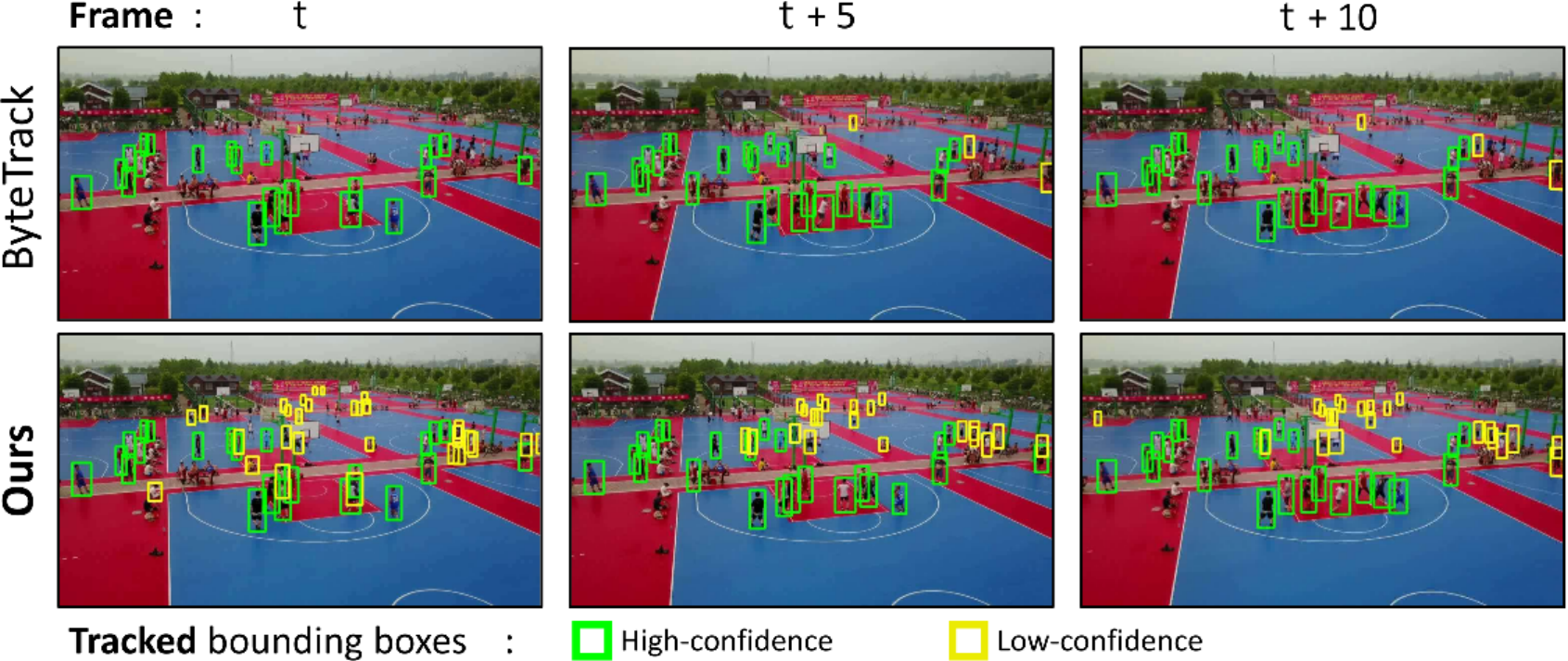}
\vspace{-0.3cm}
   \caption{
Tracking comparison on a VisDrone2019 test-dev sequence, focusing on `pedestrian' class with a far-reaching UAV view.
Bounding boxes are categorized as High ($s \geq 0.7$) and Low ($s < 0.7$) confidence.
The first row shows a tracking algorithm initiating tracks solely from high-confidence detections.
In contrast, the second row illustrates our approach, initiating tracking from both high and low-confidence detections.
This highlights our method's effectiveness in tracking small-scale objects in UAV footage,
particularly when detection confidence is low.
   }
\vspace{-0.3cm}
\label{fig:confidence_tracking}
\end{figure}

\noindent \textbf{MOT17}
We additionally evaluated our proposed method on the widely used MOT17 dataset, as detailed in Table~\ref{MOT17 val results}.
The results highlight the versatility of our approach, showcasing its effectiveness in non-UAV video environments.
Notably, when applied to two detectors, YOLOX and YOLOX Nano,
our method exhibited a substantial performance gap,
particularly with the lower-performing YOLOX Nano.
This underscores the efficacy of our method,
especially in handling increased low-confidence detections resulting from diminished detection performance, as illustrated in Figure~\ref{fig:confidence_tracking}.

\subsection{Ablation study}
Finally, we conducted an ablation study to evaluate the impact of each component in our new tracking algorithm.
To be specific, both BoTSORT and our SFTrack utilize ByteTrack as the backbone algorithm as a starting point. In this study, we thus demonstrate how our three proposed simple solutions enhance existing tracking methods.
For the experiment setting, we focused on the car class in the VisDrone2019, as it is one of the most commonly appearing objects in the dataset.
Table~\ref{VisDRONE2019 test ablation} provides
a detailed overview of the distinct impact of each component.
We initially replaced the motion compensation technique (MC) in BoTSORT with our proposed UAV-specific method that preserve aspect ratio, resulting in notable improvements, specifically a 1.4\% increase in MOTA and 3.7\% in IDF1.
Initiating tracking with low-confidence detections
further increased MOTA and IDF1 by 3.2\%.
Moreover, our association strategy for low-confidence detections,
utilizing traditional appearance matching algorithms contributed to
an additional increase of +1.7\% in MOTA and +0.2\% in IDF1.
This result indicates that traditional methods effectively handle frequent low-confidence detections at minimal extra expense.
Despite concerns about lighting and viewpoint variations,
their performance remains robust in consecutive frames of high-FPS videos.
In terms of tracking speed, integrating UAV MC has a similar impact on tracking speed as BoTSORT's MC. Both our system (row 4) and BoTSORT achieve a tracking speed of 13 FPS. Initiating tracking from low-confidence detections slightly reduces this to 12 FPS, while applying traditional appearance matching algorithms further adjusts it to 10 FPS, aligning with UAVMOT's tracking speed.
While these improvements slightly impact tracking speed,
our implementation in Python demonstrates the concept,
with potential for speed enhancements by transitioning to C.

\begin{table}[!t]
\caption{Ablation study on \textbf{car} class in \textbf{VisDrone2019} test-dev.}
\label{VisDRONE2019 test ablation}
\begin{adjustbox}{max width=\columnwidth}
\begin{tabular}{ l | c | c c | c | c | c c }
\toprule
\multirow{2}{*}{Model} & 1st Asso. & \multicolumn{2}{c|}{Motion Comp.} & Tracking Init. & 2nd Asso. & \multicolumn{2}{c}{Evaluation Metrics}\\
\cline{2-8}
\rule{0pt}{0.3cm} & Re-ID & MC &\textbf{UAV MC} & \textbf{w/ Low Det.} & \textbf{Traditional Algorithms} & MOTA$\uparrow$ & IDF1$\uparrow$ \\
\hline
\rule{0pt}{0.3cm} ByteTrack & & & & & & 58.6 & 67.7 \\ 
\rule{0pt}{0.3cm} & \usym{1F5F8} & & & & & 58.9 & 68.0 \\
\rule{0pt}{0.3cm} BoTSORT & \usym{1F5F8} & \usym{1F5F8} & & & & 58.7 & 71.2 \\
\rule{0pt}{0.3cm} & \usym{1F5F8} & & \usym{1F5F8} & & & 60.1 & 74.9 \\
\rule{0pt}{0.3cm} & \usym{1F5F8} & & \usym{1F5F8} & \usym{1F5F8} & & 63.3 & 78.1 \\
\rule{0pt}{0.3cm} \textbf{SFTrack} & \usym{1F5F8} &  & \usym{1F5F8} & \usym{1F5F8} & \usym{1F5F8} & \textbf{65.0} & \textbf{78.3} \\
\bottomrule
\end{tabular}
\end{adjustbox}
\end{table}

\section{Conclusion}
In this study, we introduced a novel association approach with three simple yet effective ideas for object tracking,
focusing on the unique challenges of UAV footage.
Our strategies for low-confidence detections effectively address tracking small-scale objects and managing UAV motions.
Our method outperforms established approaches across a variety of benchmarks,
as demonstrated on three widely recognized datasets such as VisDrone2019, UAVDT, and MOT17.
Furthermore, we identified and corrected errors in the existing annotations of the UAVDT dataset.
The enhanced version will be publicly accessible,
providing a valuable resource for more accurate benchmarking in the field.



\section*{ACKNOWLEDGMENT}
This work was supported under the framework of international cooperation
program managed by National Research Foundation of Korea (NRF-2021K1A3A1A49097955).

\bibliographystyle{IEEEtran}
\bibliography{root_cite}

\end{document}